\documentclass[10pt,twocolumn,letterpaper]{article}

\usepackage{cvpr}
\usepackage{times}
\usepackage{epsfig}
\usepackage{graphicx}
\usepackage{amsmath}
\usepackage{amssymb}
\usepackage{xcolor}
\usepackage{caption}
\usepackage{float}
\usepackage{enumitem}

\cvprfinalcopy % *** Uncomment this line for the final submission

\def\cvprPaperID{****} % *** Enter the CVPR Paper ID here

\begin{document}

%%%%%%%%% TITLE
\title{Causal Explanations of Image Misclassifications}

\author{Miles Bennett\\
Stanford School of Engineering \\
{\tt\small milesb1@stanford.edu}
\and
Yan Min\\
Stanford School of Medicine\\
{\tt\small yanmin@stanford.edu}
}

\maketitle
%\thispagestyle{empty}

%%%%%%%%% ABSTRACT
\begin{abstract}
   The causal explanation of image misclassifications is an understudied niche, which can potentially provide valuable insights in model interpretability and increase prediction accuracy. This study trains CIFAR-10 on six modern CNN architectures, including VGG16, ResNet50, GoogLeNet, DenseNet161, MobileNet\_V2, and Inception\_V3, and explores the misclassification patterns using conditional confusion matrices and misclassification networks. Two causes are identified and qualitatively distinguished: morphological similarity and non-essential information interference. The former cause is not model dependent, whereas the latter is inconsistent across all six models. 
   
   To reduce the misclassifications caused by non-essential information interference, this study erases the pixels within the bounding boxes anchored at the top 5\% pixels of the saliency map. This method first verifies the cause; then by directly modifying the cause it reduces the misclassification. Future studies will focus on quantitatively differentiating the two causes of misclassifications, generalizing the anchor-box based inference modification method to reduce misclassification, exploring the interactions of the two causes in misclassifications.
\end{abstract}

%%%%%%%%% BODY TEXT
\section{Introduction}

Convolutional Neural Network (CNN) represents one class of black-box machine learning algorithms. Despite its popularity in many applied fields, such as medical and legal settings, the lack of causal interpretability often spurs ``trust issues" and criticism from a broader research community and front-line users. Typical frameworks employed in such settings are often required to be causal and interpretable. Consider the following scenario: a CNN algorithm is deployed to predict cardiac arrest using electrocardiogram (ECG) in the outpatient of geriatric medicine. A false negative result means a cardiac arrest event is missed by the algorithm, this could lead to catastrophic outcomes for the patient. Naturally, under this circumstance, the first reaction is to ask: why does the algorithm fail to classify certain cases? Scenarios as such are commonly seen across different fields. However, an satisfying answer to this ``why" question has rarely been adequately provided. Instead, researchers often focus on striving to enhance the algorithm performance with a goal of increasing the prediction accuracy. Hence, insights from the misclassified results are often overlooked or understudied. Investigating the causal interpretability of misclassifications can first, give a direct answer to the previously mentioned ``why" question; second, estimate the attributable causal effects to the input data and the model architecture; finally, provide an opportunity to modify the potential causes based on evidence, increase prediction accuracy.\\

\section{Related Work}
Only in the recent years has black-box model interpretability gained increasing attention. Traditional interpretability frameworks can be classified into two categories: (1) Inherently interpretable models that generate explanations during the training process, such as attention networks, disentangled representation learning; (2) Post-hoc models that first generate an abstract of training concept and map the abstract onto a more interpretable domains, such as saliency maps, example-based explanations. \cite{1} Such frameworks only draw attentions onto the feature importance at its face value and its contribution to predictive accuracy, which are correlation-based considerations.\cite{2} However, causal interpretation concerns questions such as how does altering input features or a component of the algorithm architectures change the prediction results. \\

Existing causal inference frameworks can generally be classified into model-based causal inference and sample-based causal inference. The former answers questions such as ``What is the impact of the j-th filter from i-th layer?", whereas the latter focuses on explaining the causal mechanism through generated counterfactual images. \cite{3,4} Some other efforts include treatment effect framework, global explanation framework, and algorithm fairness metrics, etc. \cite{5,6,7} \\

Unlike the previously mentioned methods, our causal framework takes on a ``backward" approach by investigating the potential causes of misclassification in six modern CNN architectures: VGG16, ResNet50, GoogLeNet, DenseNet161, MobileNet\_V2, Inception\_V3; and strives to answer the following three questions: (1) Is there a generalizable (invariant) pattern of misclassifications across all six architectures? (2) Can we deduce testable causal hypothesis from the observed patterns? (3) By precisely modifying the identified causes, will the number of original misclassifications be reduced without introducing new misclassifications? The input of this study is image the output is the class label of the corresponding image. More details about the data set will be described in the method section. 

%-------------------------------------------------------------------------

\section{Method}

{\bf Obtain Misclassified Samples} We first train the image data on VGG16, ResNet50, GoogLeNet, DenseNet161, MobileNet\_V2, Inception\_V3; then classify the test set using the trained networks. Here, we briefly describe the unique features of each selected architecture. VGG16 comparing to the earlier architectures uses smaller ($3 \times 3$) convolutional filters but more layers; ResNet50 includes the residual module to enable even deeper network without information loss; GoogLetNet uses efficient ``inception" module and removes the fully-connected layer to save memory; DenseNet161 connects each layer to every other layer in the forward path to increase training efficiency; MobileNet\_V2 uses thin ``bottleneck" layers for the input and output of the residual module; finally Inception\_V3 incorporates many previously mentioned ideas into one model. The input layers for each model are modified to be compatible with CIFAR-10 images. All training is performed using Pytorch Lightning \cite{falcon2019pytorch}

A consistent set of hyperparameters are used to train each model. The goal is to achieve greater than 90\% accuracy for the CIFAR-10 dataset. The following set of hyperparameters were used for training each model:
\begin{itemize}[noitemsep]
  \item \emph{Optimizer:} Stochastic Gradient Descent (SGD)
  \item \emph{Batch Size:} 256
  \item \emph{Learning Rate:} 0.01
  \item \emph{Number of Epochs:} 100
\end{itemize}

The training and validation curves are included in \textbf{Appendix Figure \ref{fig:training}}.

After testing, we then record all the misclassified images including their correct classes and misclassified classes. Two descriptive statistics are calculated here: the class-wise missclassification rate $u_i$ and the conditional misclassification rate $v_{j|i}$, where $i \in {1, 2, ... , 10}$ is the correct image class, $j \in {1, 2, ... , 10}$ is the misclassified class. Thus we have

\begin{align} \label{eu_eqn}
    u_i = \frac{n_{i \to j, j \neq i}}{n_i} \\
    v_{j|i} = \frac{n_{j|i}}{n_{i \to j, j \neq i}}
\end{align}

where $n_i$ is the number of images in class $i$, $n_{i \to j, j \neq i}$ is the total number of images in class $i$ that are misclassified into other classes, $n_{j|i}$ is the number of images from class $i$ that are misclassified as class $j$. In other words, $u_i$ is the marginal probability that an image from class $i$ gets misclassified, $v_{j|i}$ is the conditional probability that, conditioning on the image from class $i$ is misclassified, the probability this image is classified into class $j$. 

{\bf Extract Misclassification Patterns} Using the conditional probabilities $v_i$, a 10 by 10 confusion heatmap is constructed to capture the misclassification patterns for every CNN architecture. We first qualitatively assess the types and distributions of the misclassifications; then summarize the observed patterns within each heatmap and compare across all architectures. We create intuitive misclassification network with each node representing a class and directed edges coming from the correct class to the misclassified class. Weighted in-degrees for each node are calculated as the following:  

\begin{align} \label{eu_eqn}
    d_i = \sum_{j = 1}^{10} \mathbb{I}(j \to i, j \neq i)v_j
\end{align}

Where the in-degree of class $i$ is the sum of number of other classes $j$ misclassified into $i$ weighted by the conditional probability $v_j$. The in-degrees of each node are also used in comparing misclassification patterns across all network architectures. 

{\bf Causal Hypothesis} To generate the causal hypothesis, we gather evidence from the following three aspects: 1) the misclassification results from these six selected architectures; 2) the common misclassification structures from all the heatmaps; 3) the distributions of the scores of misclassified classes and summary statistics $u_i$, $v_{j|i}$, and $d_i$. The testable causal hypothesis will focus on the following two potential causes of misclassifications: the innate inter-class distances and the non-essential information interference. The former concerns the morphological similarities across misclassified classes that could also confuse human-performed classification. The latter describes the phenomenon that the pixels outside of the target objects interfering and dominating the misclassifcations of the target objects. The first cause may potentially make up the majority of the irreducible error of misclassification and could be harder to improve upon in comparison to the second one. Besides differentiating the two causes qualitatively, this study also attempts to test the difference statistically across all selected architectures. 

Based on the premises above, causal questions we posit here are: 1) Is the innate morphological similarity consistently causing misclassifications across all selected architectures? 2) Is the non-essential information interference causing misclassifications across all selected architectures? 3) Can we qualitatively and quantitatively differentiate the two causes of misclassifications? 4) Can we correct non-essential information interference to reduce misclassifcations?

\textbf{Hypothesis Testing} The first two questions concern the following two aspects: is there a specific cause of misclassification and is this cause model dependent? Two answer these two questions, we will employ human domain knowledge of each class and incorporate evidence from the confusion heatmaps and misclassification networks. Therefore, the two causes are categorized qualitatively. We simply apply the same categorization across all six models to argue the model dependence. 

The answer to questions 1) and 2) lead to the first half of the answer to question 3). However to the qualitatively differentiate the two causes, we first qualitatively categorize the two types of misclassifications, then compute the difference of the scores between correct class and misclassified class. Concerning the relatively small sample size, we assume the score difference for each misclassification category follows a t-distribution, with the following density function:

\begin{align}
    f(s) = \frac{\Gamma(\frac{d+1}{2})}{\sqrt{d\pi}\Gamma(\frac{d}{2})}(1 + \frac{s^2}{d})^{-\frac{d+1}{2}}
\end{align}

where $d$ is the number of degrees of freedom, $\Gamma$ is the gamma function, $s$ is the random variable - score difference. Let $n_1$ and $n_2$ denote the sample sizes of the two types of misclassifications, and assume the two sets of scores have similar variance, then we may construct the t-statistics:

\begin{gather}
    t=\frac{\Bar{s}_1 - \Bar{s_2}}{SE_p\sqrt{\frac{1}{n_1} + \frac{1}{n_2}}}\\
    SE_p=\sqrt{\frac{(n_1 - 1)SD_{s_1}^2 + (n_2 - 1)SD_{s_2}^2}{n_1 + n_2 - 1}}
\end{gather}

where $SD_{s_1}^2$ and $SD_{s_2}^2$ are the standard deviations of the two score distributions, the degrees of freedom here is $n_1+n_2-1$.

To answer question 4), we first need to show the existence of non-essential information interference in the image to be classified. Then ``surgically" remove the interference, have the model to reclassify the modified image, then record the classification results to observe any decrease in misclassifications. This step is equivalent to the ``$do(x)$" in causal inference, where we modify the cause and observed the potential outcomes under both conditions before and after ``$do(x)$" is implemented. To achieve this set of goals, we first extract the saliency maps of the misclassified images caused by non-essential information interference. Based on our hypothesis, the saliency maps shall indicate peripheral regions outside of the target objects that are driving the model classifications. We then modify the original image by first selecting the top 5\% pixels on the saliency map, generate bounding boxes around the selected pixels with width and height as $\Delta x$ and $\Delta y$, set the pixels within the bonding boxes in the original image to 0, and reclassify the modified image. The driving goal here is to make the minimum image alteration to achieve the largest reduction in misclassification. The top percentages of the pixels on the saliency map, the width and height of the bounding boxes are potentially hyperparameters that can be tuned to optimize the performance of this process, however, this study will only take the first step to explore the efficacy of this idea, the detailed tuning step will be included in our future work.

{\bf Compare to Baseline Methods} We compare our training and test results of CIFAR-10 from all included networks with the state-of-the-art training and test results from the machine learning community. Since causal inference is a quite new topic in machine learning and is quickly evolving, we are not able to identify comparable baseline methods in this regard. However, we will be doing the rolling literature review, if there are comparable studies emerging, we will include the additional comparison in our final paper. 

%-------------------------------------------------------------------------

\begin{figure}[h]
\includegraphics[width=\linewidth]{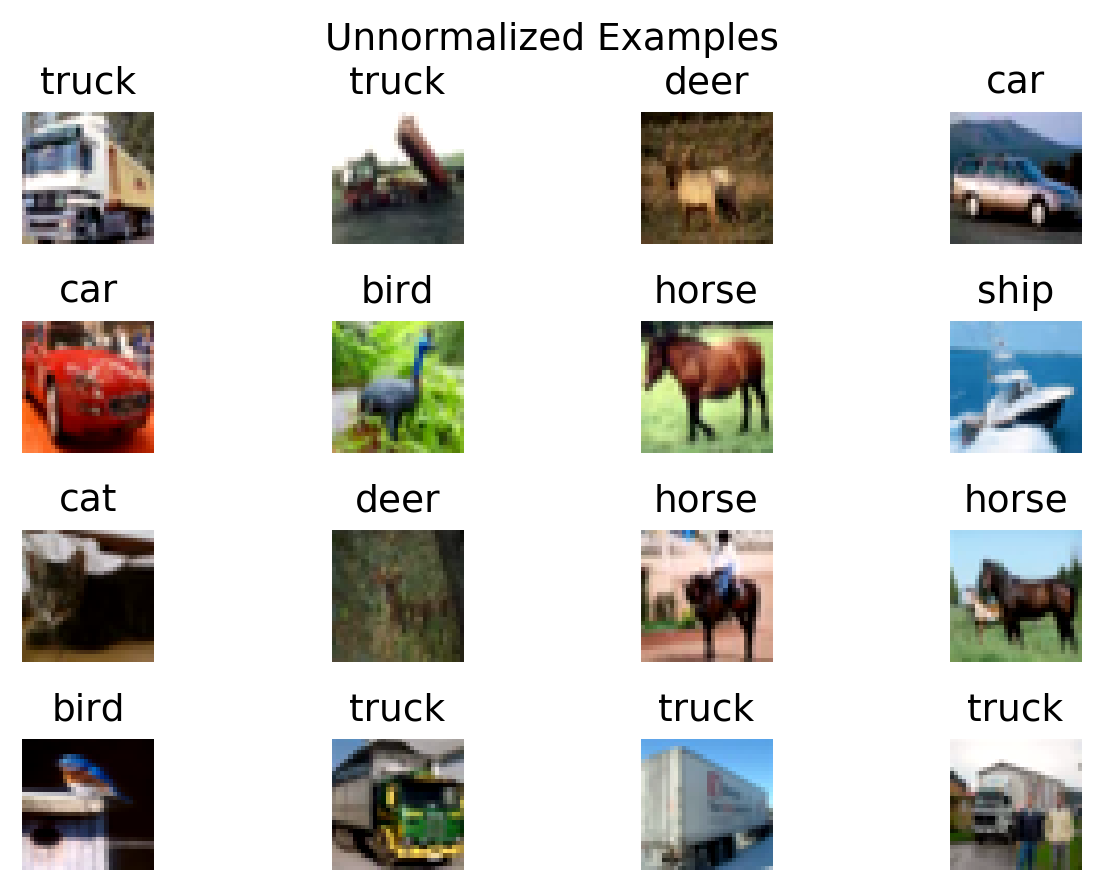}
\caption{CIFAR-10 Examples}
\label{fig:example}
\end{figure}

\section{Dataset and Features}

This study uses the CIFAR-10 dataset, which contains 60,000 32x32 color images from 10 classes (6,000 images per class). 10,000 out of the total images are held out as a test set with 1,000 test images per class. The 10 classes are planes, cars, birds, cats, deer, dogs, frogs, horses, ships, and trucks. \textbf{Figure \ref{fig:example}} includes the original sample images from the dataset. Each image is channel-wise normalized as the following: 

\begin{gather}
    \mu_{c} = \frac{1}{32 \times 32} \sum_{n=1}^{50000} \sum_{i=1}^{32}\sum_{j=1}^{32}x_{c,n,i,j}\\
    \sigma_{c}^2 = \frac{1}{32 \times 32}\sum_{n=1}^{50000} \sum_{i=1}^{32}\sum_{j=1}^{32}(x_{c,n,i,j}-\mu_c)^2\\
    x_{c,n,i,j}^{'} = \frac{x_{c,n,i,j} - \mu_c}{\sigma_c}
\end{gather}

\begin{figure}[h]
\includegraphics[width=\linewidth]{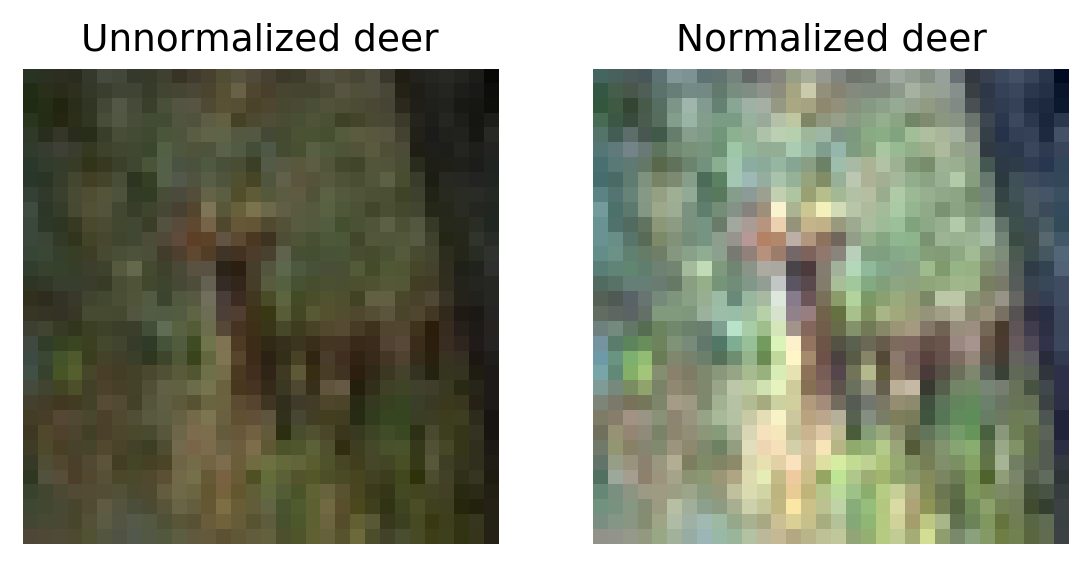}
\caption{CIFAR-10 Normalized Deer Example}
\label{fig:deerexample}
\end{figure}

Where $c \in {(r, g, b)}$ denotes the three color channels of the image. $n$ is indexing the training image. $\mu_c$ and $\sigma_c$ are the mean and standard deviation of each channel. $x_{c,i,j}$ and $x_{c,i,j}^{'}$ are the pixel value at $(i,j)$ of channel $c$, before and after normalization, respectively. \textbf{Figure \ref{fig:deerexample}} provides an example of before and after normalized.

\begin{figure}[h]
\includegraphics[width=\linewidth]{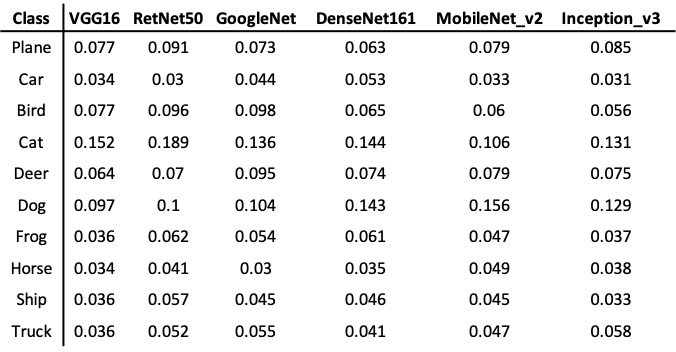}
\caption{Misclassification Rate by Class and Architectures}
\label{fig:results}
\end{figure}

\begin{figure*}[h]
\includegraphics[width=0.8\paperwidth]{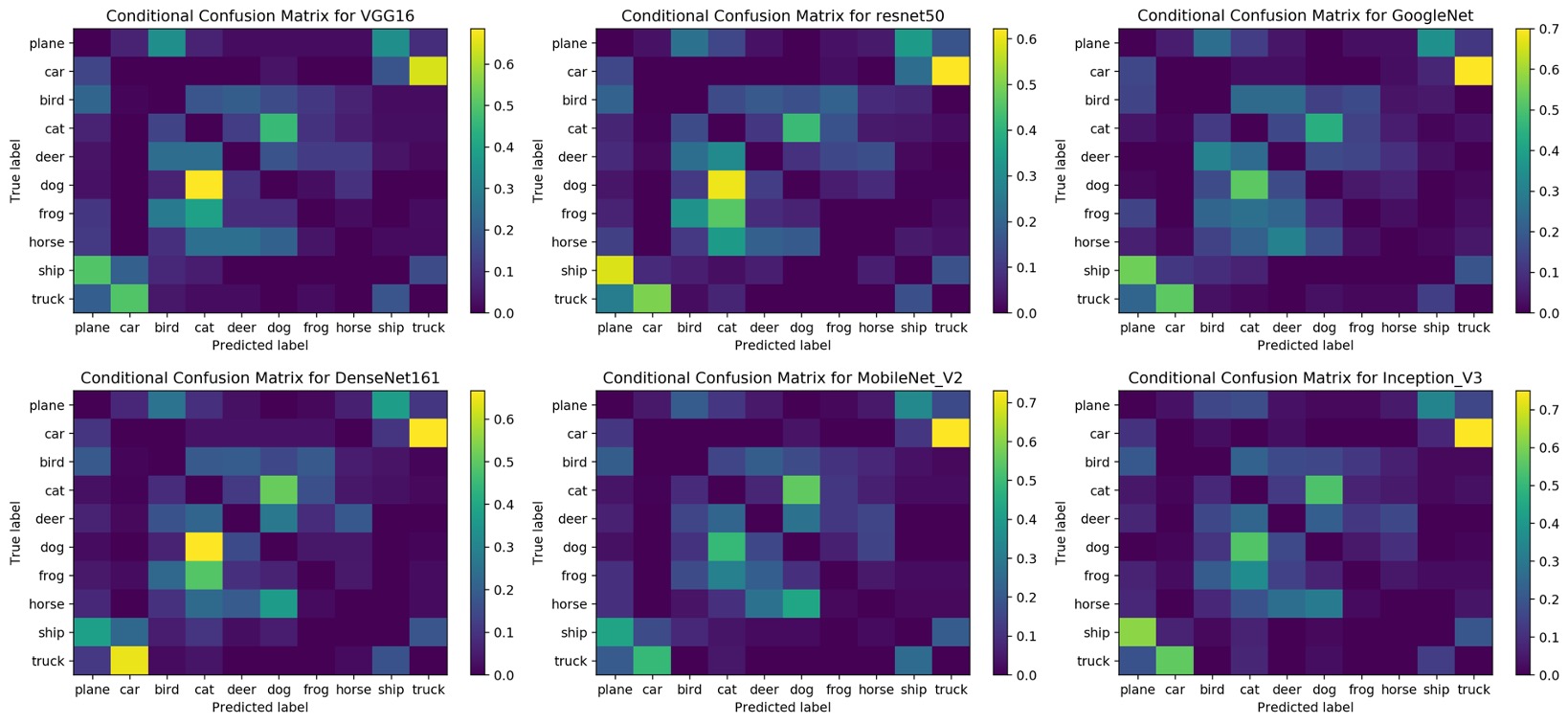}
\caption{The conditional confusion heatmaps from six architectures. Each row represents the correct class, each column represents the misclassified class. Color scale indicates the conditional misclassification rates for each class. From blue to yellow, the conditional probability increases.}
\label{fig:heatmap}
\end{figure*}

%-------------------------------------------------------------------------
\section{Results}

%------------------------------------------------------------------------

\subsection{Classification Performance Comparisons}

Papers on the state-of-the-art architectures do not conventionally report class-wise misclassification rate, as it is rarely the goal of interest, hence we will not be able to compare our self-trained class-wise misclassification rate to the baseline models. However, the architectures under investigation are comparable to state-of-the-art performance accuracy on the CIFAR-10 dataset. At the time of writing, the top performing architecture on CIFAR-10 (without extra training data) is PyramidNet, which achieved 98.5\% accuracy on the test set.\cite{8} By comparison, the top performing model in this study, Inception-v3, achieved 93.3\% on the test set. \textbf{Figure \ref{fig:results}} summarizes the misclassification rates by each class and by each trained architecture. We conducted a chi-squared heterogeneity test to examine the misclassifcation distributions across six models. The $p$ value is close to 1, which indicates the class-wise misclassification rates is homogeneous across all architectures are homogeneous. \\

%------------------------------------------------------------------------
\subsection{Misclassification Patterns}

As indicated in the homogeneity test, all six architectures share similar misclassification patterns. \textbf{Figure \ref{fig:heatmap}} presents the conditional misclassification rates by class for all six networks. An obvious pattern across all six models is the cat $\leftrightarrow$ dog misclassifications, where dogs are most likely misclassified as cats and cats are likely misclassified as dogs. Another pattern that is similar to the cat $\leftrightarrow$ dog misclassifications is the car $\leftrightarrow$ truck misclassifications. Noting these two patterns are symmetrical. In other words, two classes are ``trading" misclassified images. 
The third pattern is the plane $\leftrightarrow$ ship misclassifications. The above mentioned symmetry is only observed in ResNet50, GoogLeNet, and DenseNet161. In VGG16, MobileNet\_V2, and Inception\_V2, such symmetry does not persist. However, all these three networks have ships more likely to be classified as plane than the other way around. 

\begin{figure*}[h]
\includegraphics[width=0.8\paperwidth]{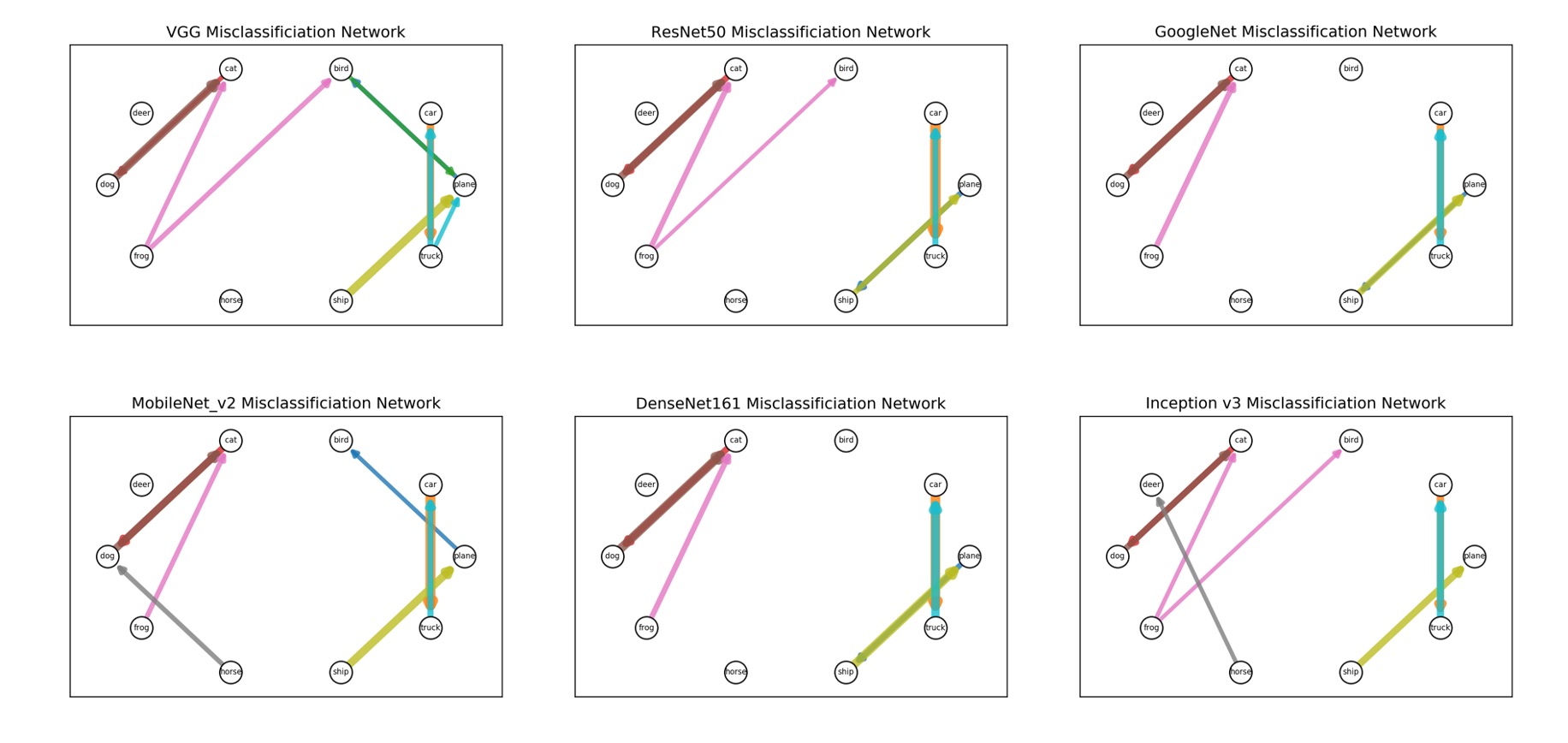}
\caption{The misclassification networks visualize the class-wise misclassifications with each node representing one class, directed edges pointing from the correct class to the misclassified class. The threshold for the directed edges is 0.3, $v_{j|i} \geq 0.3$}
\label{fig:network}
\end{figure*}

To visualize the symmetry and calculate the in-degrees ($d_i$), we constructed a misclassification network (\textbf{Figure \ref{fig:network}}) for each model with 0.3 as the threshold for the misclassification rate. From \textbf{Figure \ref{fig:network}}, we observe two universal patterns across all six models: 1) the symmetrical patterns in the cat $\leftrightarrow$ dog and car $\leftrightarrow$ truck misclassifications in all six models; 2) the asymmetrical pattern of frog $\to$ cat miscalssifcation, as all six models we observe frog is more likely to be categorized as cat than vice versa. The third pattern shown in the network is the inconssitent ship $\leftrightarrow$ plane misclassification as mentioned above. The rest of the non-universal patterns are all asymmetric. Hence we may conclude with caution that symmetric patterns are not model dependent, whereas asymmettic patterns do not have such guarantees.

\begin{figure*}[h]
\includegraphics[width=0.8\paperwidth]{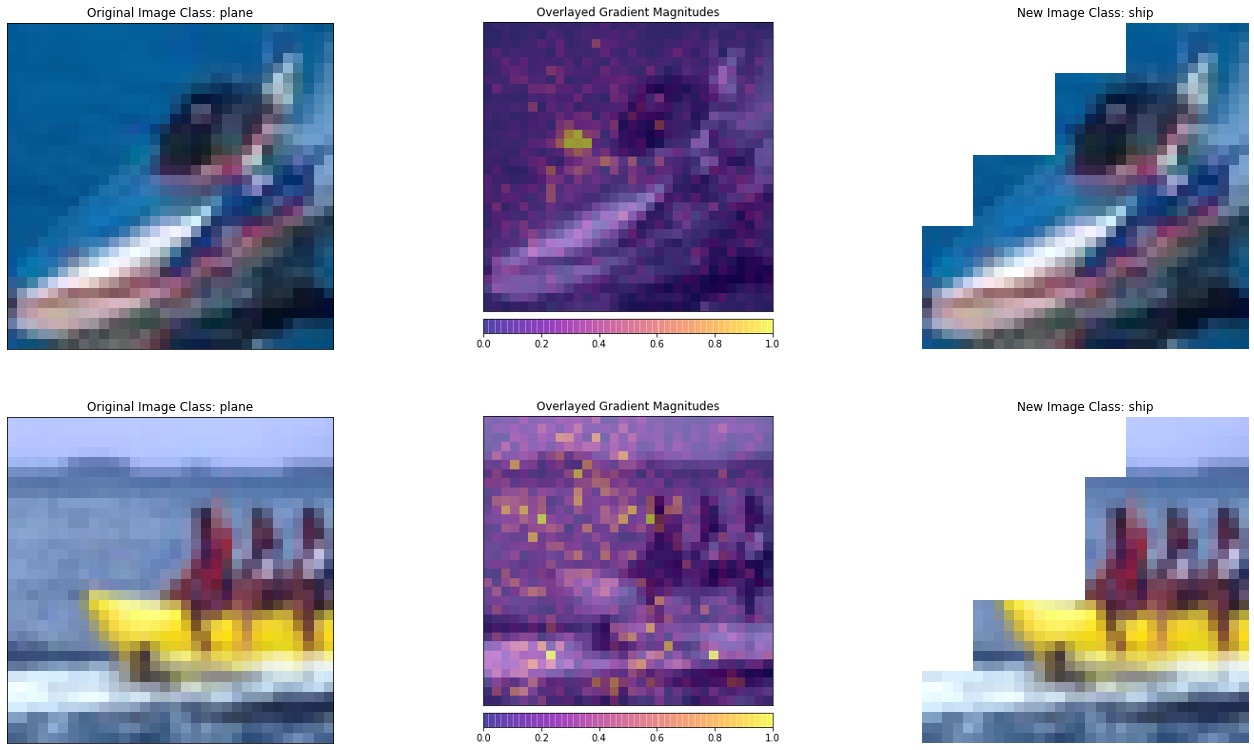}
\caption{The cause modification demo shows the the ships are originally classified as plane (the left column), the saliency maps of the classification (the middle column), and the ships are correctly classified as ships after cause modifications (the right column).}
\label{fig:saliency}
\end{figure*}

\subsection{Two Causal Hypothesis and Testing}
Based on the evidence presented previously, we make the following two hypothesis: 1) the two symmetric patterns, cat $\leftrightarrow$ dog, car $\leftrightarrow$ truck, are potentially caused by innate morphological similarities of the two classes; 2) the asymmetric pattern, frog $\to$ cat and ship $\to$ plane, allows us to hypothesize that the misclassifications between these two classes are potentially caused by non-essential information interference.

To explore the two causes of misclassifications, we have compared the distributions of the score ratios (correct class / misclassified class) of the two types of misclassifications (morphology vs interference). The t-tests result are not significant, as the score ratio distributions are similar for these two categories. We include histograms for the score ratio distributions in \textbf{Appendix Figure \ref{fig:distribution}}. 

We thereby conclude that although the two types of causes are easily differentiated qualitatively, the metrics chosen in this study cannot differentiate the two causes of misclassifications quantitatively.

\subsection{Cause Modification and Reclassification}
Due to the time constraint, we are only able to conduct a cause modification experiment using VGG16. In \textbf{Figure \ref{fig:saliency}}, we provide two demo examples of this experiment. As stated previously, we hypothesize that the ship $to$ plane misclassification is caused by non-essential information interference. Hence, the two examples are both from the ship images that are initially classified as planes. To verify this hypothesis of interference causes misclassification, we first generate the saliency map for the two demo images \cite{captum2019github}. The saliency maps in the middle column indicate the image classification is heavily based on the background of the image instead of the target object, which is the ship in these two examples. Hence, we implement the method proposed in the \textbf{Method} section to construct bonding boxes around the top 5\% of the pixels of the saliency map. Top 5\% of these pixels that belong to the target object are spared from this operation. Then we set the pixels within the bonding boxes to 0 on the original images, and rerun the VGG16 network to reclassify the images. We are able to obtain the correct classification based on this operation. The idea here is similar to image segmentation and background removal, however we do not need perfect segmentation to achieve the correct classification. Furthermore, segmentation requires nontrivial computation, which is time consuming and computationally expensive. The driving goal here is to implement minimum modification to obtain the correct classification. 

%-------------------------------------------------------------------------
\section{Conclusion}

This study systematically categorizes the causes of misclassifications into morphological similarity and non-essential information interference. The two categories of misclassifications are observed across all six selected architectures. Although it is intuitive to qualitatively differentiate the two causes, the metrics chosen in this study is not able to provide quantitative differentiation. Interestingly, morphological similarity caused misclassifications tend to be symmetric and not model dependent, where as non-essential inference caused misclassifications tnd to be inconsistent regarding model dependency.

The saliency map is able to verify the non-essential information interference caused misclassifications. Directly modifying the cause shows success in decreasing the misclassification rate. The size of the anchor box and the top percentage of the saliency map pixels are considered as hyper parameters. The goal is to make the smallest change to the original image and maximize the reduction in misclassification.

The future direction for this study includes choosing an appropriate metric to quantitatively differentiate the misclassifications due to different causes, explore the generalizability of the proposed cause-modification method to further reduce the non-essential information interference caused misclassifications, investigate to what extend the morphological similarity caused misclassification is reducible, and explore the potential interaction between the two causes.
%------------------------------------------------------------------------
\section*{Appendices}

\begin{figure}[h]
\includegraphics[width=\linewidth]{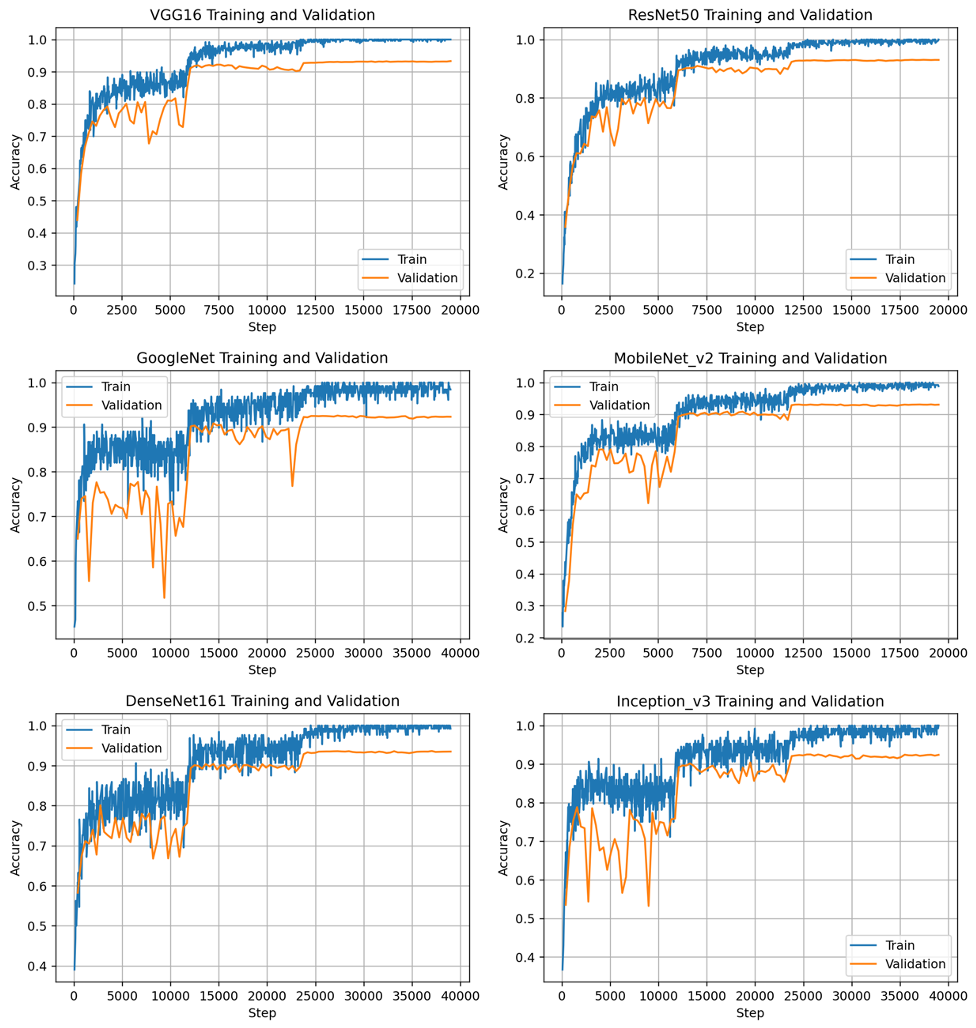}
\caption{Training and validation curves for all selected models.}
\label{fig:training}
\end{figure}

\begin{figure}[h]
\includegraphics[width=\linewidth]{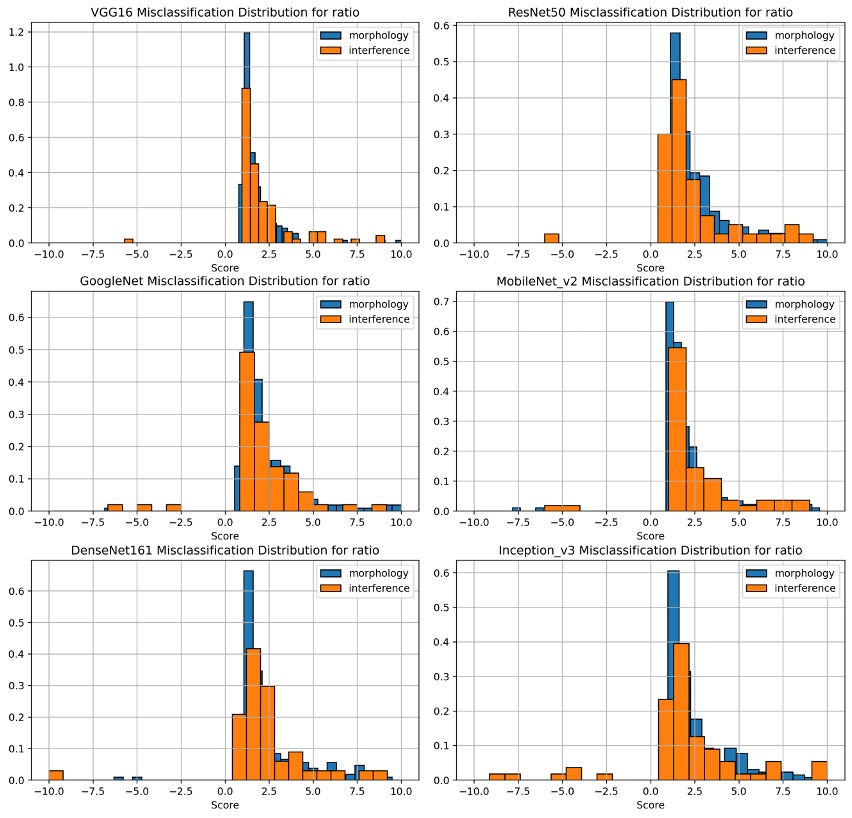}
\caption{Score ratio distributions for all selected models.}
\label{fig:distribution}
\end{figure}

%------------------------------------------------------------------------
\section*{Acknowledgements}
We sincerely thank CS231N teaching staff, especially Yi Wen who has provided constructive feedback to our project. The project has been conducted amid COVID-19 Pandemic and Nation-wide Black Lives Matter Movements, we send our condolences to the victims of these crisis and thank the Stanford community to uphold tremendous solidarity in such difficult times. 

The two author contributed equally to the project. Authors do not declare any conflict of interest.
%------------------------------------------------------------------------

{\small
\bibliographystyle{ieee_fullname}
\bibliography{egbib}
}

\end{document}